\PassOptionsToPackage{dvipsnames}{xcolor}
\documentclass[conference]{IEEEtran}
\IEEEoverridecommandlockouts
\usepackage{cite}
\usepackage{amsmath,amssymb,amsfonts}
\usepackage{algorithmic}
\usepackage{graphicx}
\usepackage{subcaption}
\usepackage[ruled,vlined, linesnumbered]{algorithm2e}
\usepackage{listings}
\usepackage{todonotes}

\usepackage{textcomp}
\usepackage{tikz}
\usepackage{url}
\usetikzlibrary{shapes.geometric,arrows.meta, positioning, calc}

\usepackage{amsthm}
\theoremstyle{plain}

\theoremstyle{definition}
\newtheorem{defn}{Definition}

\def\BibTeX{{\rm B\kern-.05em{\sc i\kern-.025em b}\kern-.08em
    T\kern-.1667em\lower.7ex\hbox{E}\kern-.125emX}}

\usepackage{soul}
\definecolor{red}{RGB}{255,0,0}
\definecolor{lightgray}{gray}{0.95}
\sethlcolor{red}

\usepackage{xcolor} 
\definecolor{LightGray}{rgb}{0.97,0.97,0.97}
\lstdefinelanguage{SPARQL}{
  basicstyle=\small\ttfamily,
  backgroundcolor=\color{LightGray},
  columns=fullflexible,
  breaklines=false,
  sensitive=true,
  frame=bt,
  aboveskip=1em,
  belowskip=1em,
  xleftmargin=.5em,
  xrightmargin=.5em,
  framexleftmargin=.5em,
  framextopmargin=.5em,
  framexbottommargin=.5em,
  framexrightmargin=.5em,
  tabsize = 2,
  showstringspaces=false,
  morecomment=[l][\color{gray}]{\#},       
  morecomment=[n][\color{blue}]{<http}{>}, 
  morestring=[b][\color{OliveGreen}]{\"},  
  keywordsprefix=?,
  classoffset=0,
  keywordstyle=\color{Sepia},
  morekeywords={},
  classoffset=1,
  keywordstyle=\color{Purple},
  morekeywords={rdf,rdfs,owl,xsd,purl},
  classoffset=2,
  keywordstyle=\color{MidnightBlue},
  morekeywords={
    SELECT,CONSTRUCT,DESCRIBE,ASK,WHERE,FROM,NAMED,PREFIX,BASE,OPTIONAL,
    FILTER,GRAPH,LIMIT,OFFSET,SERVICE,UNION,EXISTS,NOT,BINDINGS,MINUS,a
  }
}

\begin{document}

\title{Generation of skill-specific maps from graph world models for robotic systems}

\author{\IEEEauthorblockN{ Koen de Vos\textsuperscript{1}, Gijs van den Brandt\textsuperscript{1}, Jordy Senden\textsuperscript{1}, Pieter Pauwels\textsuperscript{2}, René van de Molengraft\textsuperscript{1} and Elena Torta\textsuperscript{1}}
\IEEEauthorblockA{\textit{ \textsuperscript{1}Department of Mechanical Engineering , \textsuperscript{2}Department of the Built Environment} \\
\textit{Eindhoven University of Technology}}}

\maketitle

\begin{abstract}
With the increase in the availability of Building Information Models (BIM) and (semi-) automatic tools to generate BIM from point clouds, we propose a world model architecture and algorithms to allow the use of the semantic and geometric knowledge encoded within these models to generate maps for robot localization and navigation. 
When heterogeneous robots are deployed within an environment, maps obtained from classical SLAM approaches might not be shared between all agents within a team of robots, e.g. due to a mismatch in sensor type, or a difference in physical robot dimensions. Our approach extracts the 3D geometry and semantic description of building elements (e.g. material, element type, color) from BIM, and represents this knowledge in a graph. Based on queries on the graph and knowledge of the skills of the robot, we can generate skill-specific maps that can be used during the execution of localization or navigation tasks. The approach is validated with data from complex build environments and integrated into existing navigation frameworks.  
\end{abstract}

\begin{IEEEkeywords}
Robot navigation, Building Information Models (BIM), Graph World Models, Navigation, Localization.
\end{IEEEkeywords}

\section{Introduction}
There is an increasing demand for task automation using robotic systems in different sectors such as logistics~\cite{farinelli2017advanced} and agriculture~\cite{fountas2020agricultural}. The challenge in these environments is to deal with unpredictable situations, caused by a lack of geometric structure, variations in objects and unpredictable behavior of an actor, such as humans, animals or other robots. In order to let the robot understand how well it is performing its task, it needs to have knowledge about the context in which it is deployed. This knowledge about the task requirements, environment and robot, should be described in what this work calls a world model. For common robot-skills, like navigation and localization, a world model could be a geometric map of the environment. In classical SLAM approaches such a map is created directly from sensor data, which makes it hard to share with other robots, that might possess different sensors and (motion) skills. This calls for a representation of environmental knowledge on a higher semantic level than the data-level, such that it is agnostic to the sensor and it can be shared between several heterogeneous robots. 

Other typical requirements posed to large-scale robotic systems relate to their ease of deployment and possibilities for semi-automatic reconfiguration.
Research shows that these requirements can be addressed by designing knowledge bases shared by the heterogeneous robots which allow information sharing between them as well as the automatic re-configuration of skills and maps (e.g.~\cite{tenorth2017representations, waibel2011roboearth}).
In this paper, we propose a novel approach to designing knowledge bases, in the form of graph world models, that can be shared by heterogeneous robots and from which robot-specific views can be generated. 

\begin{figure}
  \centering
  \includegraphics[width=\columnwidth]{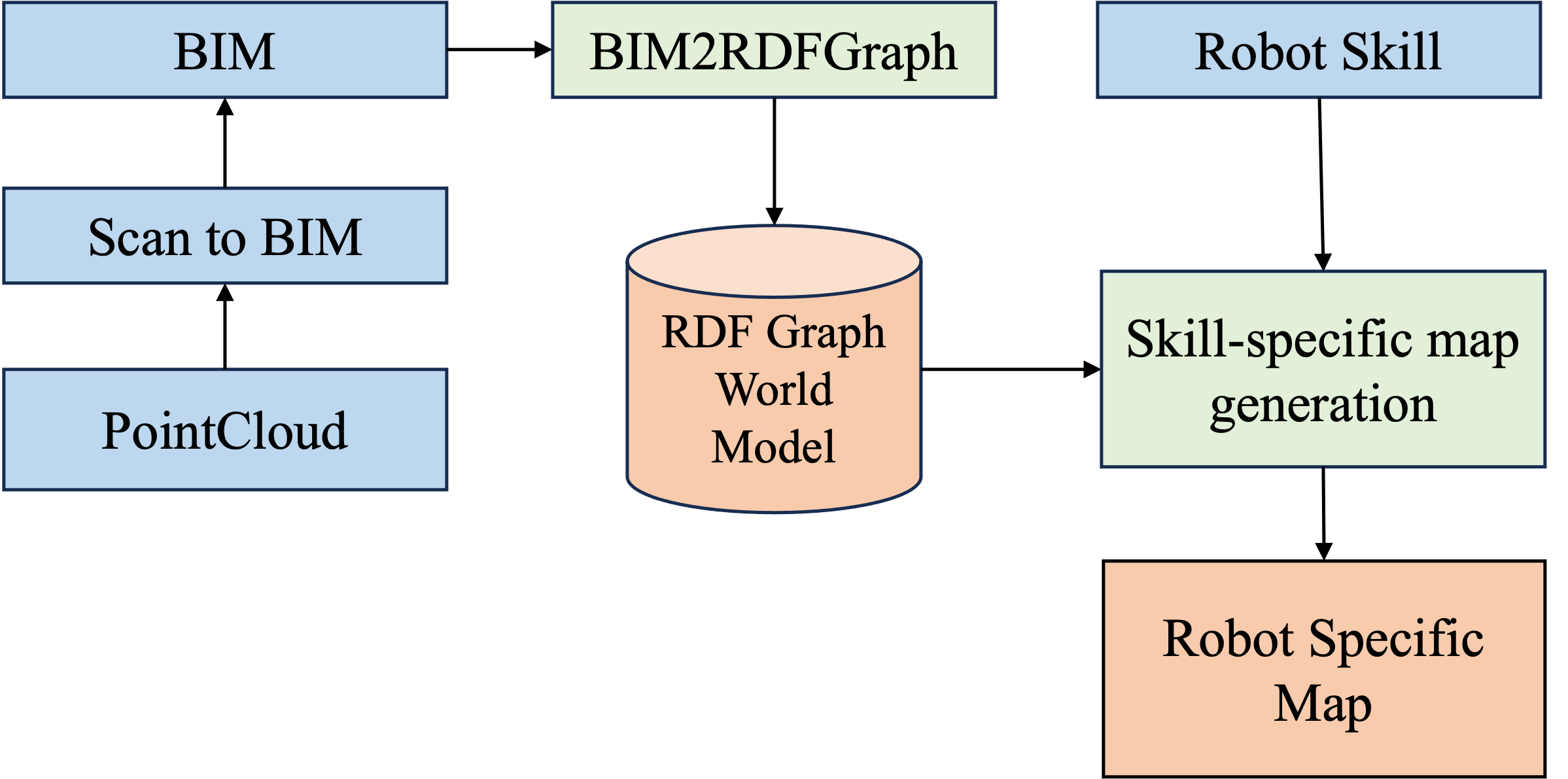}
  \caption{Conceptual architecture. Blue: provided input. Green: novel algorithms. Orange: algorithm's output. }
  \label{fig:concept}
\end{figure}

The proposed graph world model represents the building elements of an environment with associated geometry and semantic properties. By using the semantics to query the specific building elements and by interpreting the geometry based on the specifics of the robot, we demonstrate that it is possible to automatically generate robot-specific views (e.g. tailored to the type or height of its sensors, or the physical dimensions of the robot itself) in the form of environmental maps which can be integrated in navigation and localization frameworks. 
We particularly focus on ease of deployment addressing the latter from two perspectives: the automatic initialization of the graph world models from Building Information Models (BIM) and the automatic extraction of robot-specific maps from the shared world model.
The graph is initialized by interpreting data from  Building Information Models (BIM) exported in the Industry Foundation Classes (IFC) format~\cite{IFC}. Building on our previous work~\cite{de2021queries, hendrikx2021connecting, pauwels2023live}, the semantics and the geometry of the building elements are extracted and their relations are represented in the topology of an RDF graph. The use of BIM and IFC is relevant to industrial facilities such as warehouses for which these models are normally available and used throughout the design of the logistic process~\cite{he2021bim}. If a BIM model is not available, it is still possible to create one by collecting point-cloud scans of the environment \cite{BOSCHE2010107,BOSCHE2015201}. The point cloud is subsequently segmented into semantic objects of interest, annotated and converted to a BIM model. This is a common procedure in the architectural and infrastructure domain (e.g.,~\cite{soilan20203d, qu2015usage}). Since both initialization paths result in a BIM model, the same algorithm for the construction of the graph world model can be applied. 
Second, we provide methods to generate skill-specific environmental representations derived from the graph world model focusing on robots using 2D LIDAR for navigation and localization. We show that the map generation process can be tailored to the specific skills and physical properties of the robot by only representing those elements in the world model which are selected based on the properties of the specific skill and robot.
The concept presented in the paper is displayed in Fig.~\ref{fig:concept}.
The methods are validated considering graph world models of a large university building as well as a simple lab environment.
We demonstrate the feasibility of integrating the generated maps into existing navigation algorithms by enabling a small robot to autonomously navigate a lab environment for which a graph world model is available. 
The remainder of the paper is organized as follows. Section~\ref{sec:rel_work} positions this work within the state-of-the-art. Section~\ref{sec:graphWorldModel} provides a formal definition of the graph's concepts and topology and introduces methods for its automatic initialization from BIM/IFC models. Section~\ref{sec:map_generation} introduces the queries and the methods to extract the geometry of identified semantic elements and to create navigation and localization maps tailored to the skills of the robot requesting them. The results demonstrate the potential for the integration of the generated maps into existing navigation frameworks. Finally, Section~\ref{sec:DiscAndConcl} provides an interpretation of the results discussing the potential and limitations of the presented methods, and outlines future research directions.

\section{Related Work}\label{sec:rel_work}
The idea of generating skill-specific action recipes that robots could share has received widespread attention in the robotics community. Notable is the effort by the EU-FP7 RoboEarth project \cite{waibel2011roboearth, riazuelo2015roboearth} that addressed the challenges of creating a system to describe action recipes generically such that robots with different skills could execute them with minimal changes. RoboEarth particularly focused on enabling the exchange of knowledge between robots using different ontologies or forms of recipes representations such as  OWL or Linked Data principles.  Projects related to RoboEarth were the KnowRob and KnowRob2 systems\cite{tenorth2013knowrob, beetz2018know} which provide ontologies for fine-grained action descriptions and reasoning engines to generate action recipes that are tailored to robots with different skills. Central to RoboEarth, KnowRob, and related projects is the concept of a World Model \cite{Sakagami2023, Bruyninckx2021} defined as \lq\lq \textit{the information the robot has about the world around it, and that needs to be shared between several activities.}\rq\rq \cite{Bruyninckx2021}.
This paper builds on the World Model concept, emphasizing the development of a graph that represents environmental data independently of any specific robot. This allows for the creation of custom maps suited to the abilities of the particular robot using them for online decision making.
 Contrary to RoboEarth and KnowRob we do not focus on action recipes but rather on generic environmental descriptions. We particularly address aspects related to ease of deployment by providing methods for the automatic initialization of the graph World Model. We do so by leveraging the data represented in Building Information Models (BIM) of such facilities for the initialization of the graph.
The graph World Model is created following a Linked Data approach~\cite{pauwels2021knowledge} and describes the environmental knowledge in a Resource Description Framework (RDF) graph \cite{olivares2019review}.

Recent years have seen an increasing effort to link the domain of digital construction, from which BIM originates, to the domain of robotics. Much has been done in the direction of making the data reported in Building Information Models (BIM) available to robots for task execution. For example, in our previous work, we analyzed several data flow options to extract data from BIM to environmental descriptions usable by robots \cite{pauwels2023live} and validated one of them for 2D Lidar-based semantic localization \cite{hendrikx2021connecting}. The use of BIM data for localization based on 3D lidar data has also been explored in \cite{yin2023semantic, torres2023bim}. Other approaches proposed the translation of information extracted from BIM  to URDF models of single environmental elements which are later used for designing and validating action planning algorithms\cite{kim2022bim}.
Whereas prior work has focused on the creation of maps that are specific to a given navigation or localization method, this paper focuses on generality instead. First, data from BIM are extracted and stored in an RDF graph database that represents knowledge of the environment independently on the robot or the specific navigation or localization method. Second, the RDF graph world model can be queried for specific information and maps can be generated that are tailored to the skill of the robot requesting them.
BIM is therefore regarded as an input for the automatic initialization of the graph world model. The graph world model is the live representation of the facility with a bi-directional data connection with the robots and thus can evolve over time ensuring that the world model remains in sync with changes in the real environment.
To address scenarios in which existing BIM models are not available, alternative initialization paths can be considered that rely on the semantic segmentation of 3D point clouds collected in a facility and their semi-automatic conversion to BIM models.  This procedure is common in the architectural domain ~\cite{soilan20203d, qu2015usage, perez2021scan2bim} and it is referred to as the scan-to-BIM method.

\section{Graph World Model}\label{sec:graphWorldModel}
\begin{figure}
  \centering
  \includegraphics[width=.75 \columnwidth]{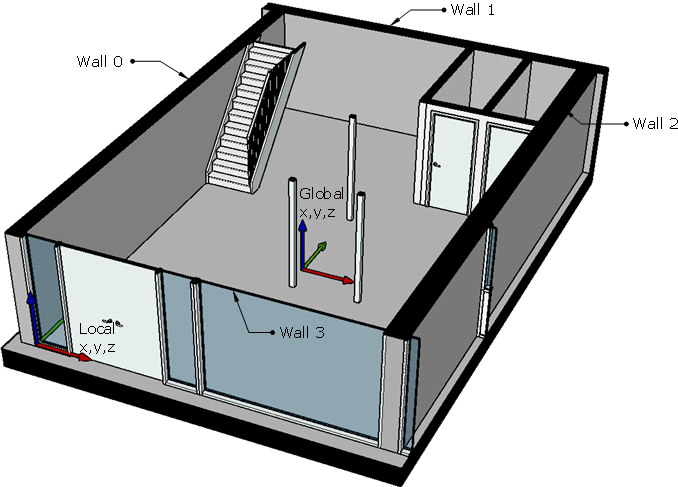}
  \caption{Representation of the mesh for Wall1 and positioning of the global and local reference frames.}
  \label{fig:ConceptWithMesh}
\end{figure}

To generate skill-specific environmental representations we first propose the creation of a graph world model, which we regard as a building digital twin, that is independent of a specific  robot's view of the world (e.g., sensors' height or type which results in a robot's specific environmental representation). 
In our approach, the graph world model $G=\{V,E\}$ describes a discrete set of labeled semantic elements that constitute the nodes $v_i \in V$.   Relations between nodes $(v_i, e_f, v_j)$ are described at the semantic level,  for example, edges $e_f\in E$ might refer to spatial relations such as   \lq\lq\textit{adjacent to}\rq\rq  or to properties such as 3D geometry or material type. For example, in the environment in Figure~\ref{fig:simpleenviroment_physical}, each semantic element of type wall is \textit{adjacent to} the semantic element of type space which represents the floor.  We encode the exact metric relation between the nodes of the graph through their 3D geometric shapes represented as a mesh. A mesh is defined as:
\begin{defn}[Mesh]\label{defn:Mesh}
 A tuple $M=(V_m, E_m, F_m, T_m)$ where:
 \begin{itemize}
 \item[] $V_m$ is the set of vertices where each vertex \( v_i \) in \( V_m \) is a point in three-dimensional space represented by its coordinates \( (x_i, y_i, z_i) \).
 \item[] \( E_m \) is the set of edges, where each edge \( e_j \) in \( E \) is defined as an ordered pair \( (v_a, v_b) \) of vertices. This implies that an edge is a line segment connecting two vertices.
 \item[] \( F_m \) is the set of faces, where each face \( f_k \) in \( F \) is defined as a closed loop of edges \( \{e_1, e_2, ..., e_n\} \).
 \item[] \( T_m \) is an homogeneous transformation matrix that represents the pose of the local mesh reference frame with respect to a designated global coordinate system.
 \end{itemize}
 \end{defn}
Meshes can be defined in CAD/BIM modelling environments or derived from point-cloud scans of physical entities~\cite{jakob2015instant}.
A visualization of a small room with multiple wall elements, the local coordinate systems of their geometry and the global coordinate frame are reported in Figure~\ref{fig:ConceptWithMesh}.

The graph world model is stored in a graph database and queries can be defined to extract the information that is relevant to generate a skill-specific environmental representation. In fact,  the graph world model is robot-independent hence some information represented in the graph will be relevant to a specific robot while other information will not be relevant for that robot. 
For example, a robot equipped with a 2D lidar will need to know the expected 2D contour of the elements of the environment it can perceive. The full 3D shape of the elements is superfluous. Object properties such as material or color could be relevant since they can influence the measurement quality of the LIDAR \cite{bolkas2018effect, tibebu2021lidar}. 
Differentiation between structural elements (e.g., walls, windows) and dynamic ones( e.g., furniture, humans) can be made by assigning this knowledge of those elements as a property to the nodes of the graph representing those elements.

\subsection{Graph World Model: implementation and initialization}
Since the graph world model describes the environment from a robot-independent perspective, we refer to recent results from the built environment domain \cite{pauwels2023live, baken2020linked, pauwels2021knowledge} and propose to describe the graph world model as an RDF graph (Resource Description Framework). Such RDF graphs are rather commonly used in several recent projects to describe the semantics of a building in relation to its geometric and telemetric data. These graphs are commonly referred to as linked building data (LBD) graphs, and follow corresponding LBD vocabularies. The main motivation of using these RDF technologies, as reported in \cite{PAUWELS2017145}, are (1) the ease in connecting multiple datasets using a web-based URI mechanism, and (2) the logical basis in OWL that allows reasoning according to Description Logic (DL) principles.
In \cite{pauwels2021knowledge, pauwels2023live}, buildings are composed of stories, stories are composed of spaces, and spaces contain elements, some elements can be composed from sub-elements. All entities can be assigned properties, and relations between entities can be modeled.\\

\begin{figure}
    \centering
    \begin{subfigure}{0.5\linewidth}
    \centering
        \includegraphics[width=\linewidth]{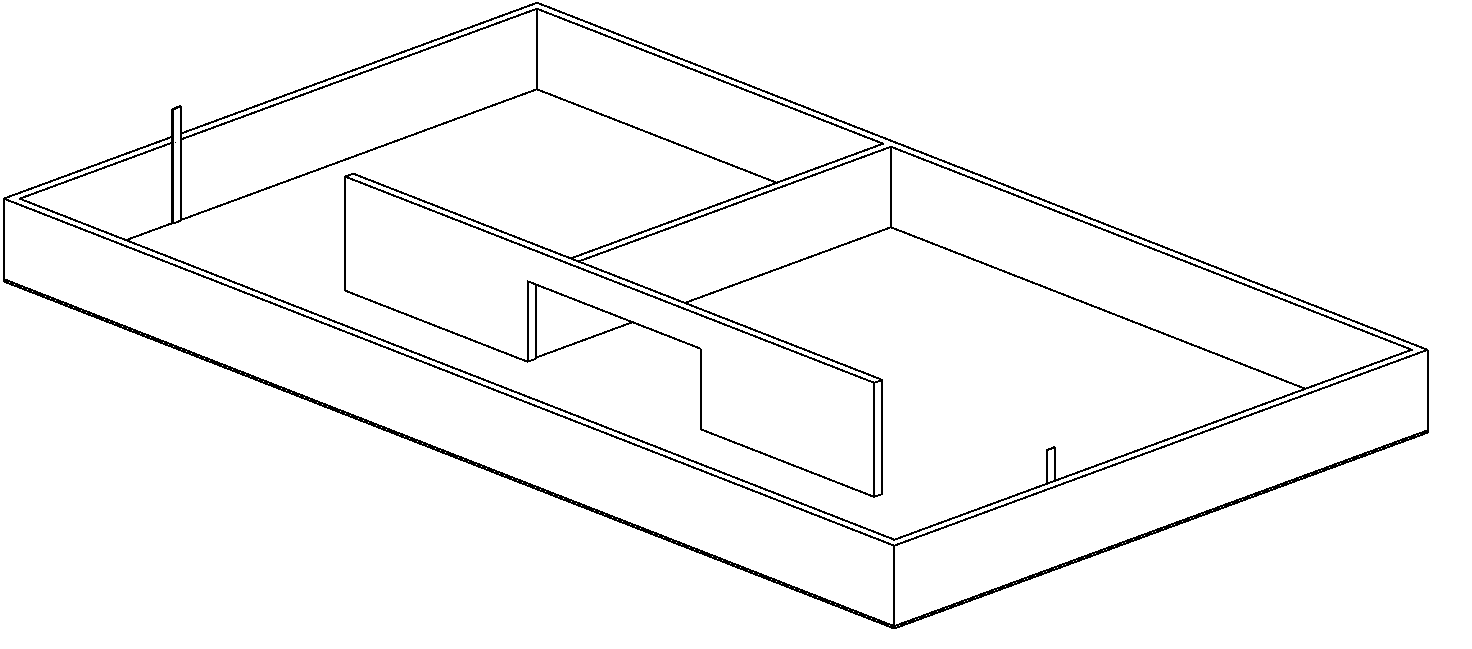} 
    \end{subfigure}%
    ~
    \begin{subfigure}{0.45\linewidth}
    \centering
        \includegraphics[width=\linewidth]{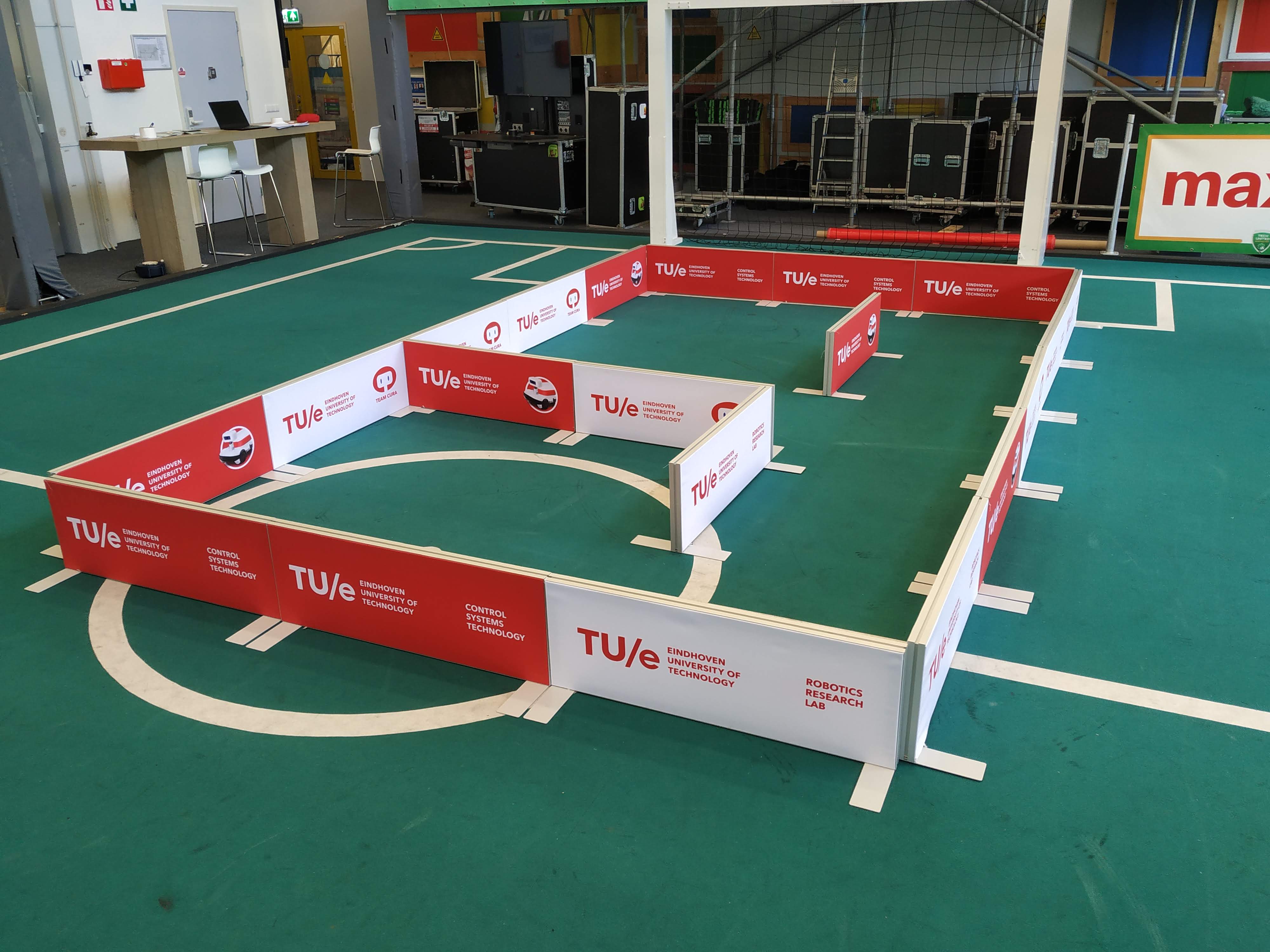} 
    \end{subfigure}%
    \caption{Render of the BIM model and physical setup of the simplified lab environment}
    \label{fig:simpleenviroment_physical}%
\end{figure}

An RDF graph  $G=(V,E)$ is a directed graph that represents relations between nodes as triples \textit{subject, predicate, object}\cite{Manola2004}. Subjects and objects are elements of the set of nodes $G$ while the predicates represent the direct edge, $e \in E$, that relates the subject to the object. In our approach, structural elements and their properties (e.g., material type, 3D geometry) constitute the nodes $v_i \in G$ of the RDF graph. Their semantic relation is described by the predicates, $e_i \in E$.  
The definition of the set of nodes $G$ depends on the available information and it can be easily expanded to accommodate new information when available as long as the RDF triples can be defined. This is particularly relevant when dynamic elements, such as humans or obstacles, need to be included in the graph world model.
To give an example, a simple environment such as the one in Figure~\ref{fig:simpleenviroment_physical} can be described by a set of nodes $G$ that include all semantic elements such as the walls and the floor, with their 3D geometry represented as a mesh (see Def.~\ref{defn:Mesh}) and properties such as color or material type represented as predicates of the tuple.

The implementation of the graph world model in terms of RDF triples that describes the environment in Figure~\ref{fig:simpleenviroment_physical} is available in the accompanying code repository \footnote{
\url{https://gitlab.tue.nl/et_projects/skillspecificmaps_from_graphworldmodels}}. The pseudo-code to describe the triples relating the area coded as \textit{Space0} to some of its adjacent walls (i.e., \textit{Wall1} and \textit{Wall2}) is reported in Listing~\ref{lst:rdft}. 
\begin{lstlisting}[language=SPARQL, caption={The triples that constitute the graph representing the environment depicted in Fig. \ref{fig:ConceptWithMesh}.}, label=lst:rdft]
#Define the relations
(Space, adjacentElement, Wall1) 
(Wall1, hasGeometry, Mesh 1)
(Wall1, hasMaterial, Concrete)
(Wall1, isStatic, True)
(Space, adjacentElement, Wall2)
(Wall2, hasGeometry, Mesh 2)
(Wall2, hasMaterial, Glass)
(Wall2, isStatic, True)
\end{lstlisting}

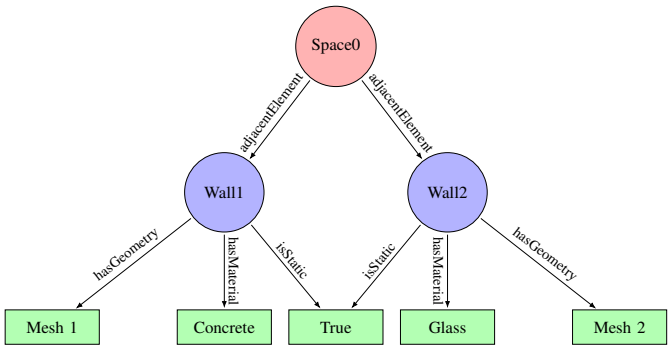
\begin{figure}
  \centering
  \resizebox{\linewidth}{!}{\begin{tikzpicture}[
    node distance=2.25cm,
    box/.style={draw, circle, text width=2cm, align=center, font=\Large, fill=#1},
    greenbox/.style={draw, rectangle, text width=2.5cm, minimum height=1cm, align=center, font=\Large, fill=#1},
    arrow/.style={-Latex, shorten >=2pt, shorten <=2pt, font=\large},
]

\node[box=red!30] (id0) {Space0};
\node[box=blue!30, below left=of id0, yshift=-1cm] (id1) {Wall1};
\node[box=blue!30, below right=of id0, yshift=-1cm] (id2) {Wall2};
\node[greenbox=green!30, below=of id1] (id6) {Concrete};
\node[greenbox=green!30, below=of id2] (id7) {Glass};
\node[greenbox=green!30, at=($(id6)!0.5!(id7)$)] (id3) {True}; 
\node[greenbox=green!30, below=of id1, left=of id6] (id4) {Mesh 1};
\node[greenbox=green!30, below=of id2, right=of id7] (id5) {Mesh 2};

\draw[arrow] (id0) -- node[above, sloped]{adjacentElement} (id1);
\draw[arrow] (id0) -- node[above, sloped]{adjacentElement} (id2);
\draw[arrow] (id1) -- node[above, sloped]{hasMaterial} (id6);
\draw[arrow] (id1) -- node[above, sloped]{hasGeometry} (id4);
\draw[arrow] (id2) -- node[below, sloped]{hasMaterial} (id7);
\draw[arrow] (id2) -- node[above, sloped]{hasGeometry} (id5);
\draw[arrow] (id1) -- node[above, sloped]{isStatic} (id3);
\draw[arrow] (id2) -- node[above, sloped]{isStatic} (id3);

\end{tikzpicture}}
  \caption{RDF graph representing some of the adjacent elements to the Space. Red: node representing a space. Blue: nodes representing adjacent elements to the space. Green nodes representing properties of the elements.}
  \label{fig:RDFGraph}
\end{figure}

A visualization of the resulting graph world model is displayed in Figure~\ref{fig:RDFGraph}. The properties of the walls are represented as green nodes. By querying all nodes connected to \textit{Space} and retrieving their meshes one can derive a robot-specific geometric representation of all known elements in the space. Specific algorithms for the latter are reported in Section~\ref{sec:map_generation}. An example SPARQL query to retrieve the meshes of the elements which are not made of glass and of certain semantic type is reported in Listing~\ref{lst:SPARQL}.   

\begin{lstlisting}[language=SPARQL, caption={SPARQL query to retrieve the meshes of all elements which are not made of glass, and of a certain set of semantic types.}, label=lst:SPARQL]
SELECT ?t ?f ?v WHERE { 
    ?s a bot:Element.
    ?s props:T ?t.     # Hom. Transf.
    ?s props:Faces ?f. # Faces of Mesh
    ?s props:Verts ?v. # Vertices of Mesh
    FILTER(?type in (beo:Wall, beo:Column))
    MINUS{?s props:Material props:Glass}}
    \end{lstlisting}


\subsection{Graph initialization}
For complex infrastructures, such as airports, warehouses, hospitals, etc., the creation of graph world models can become a time-consuming and error-prone task.
3D BIM modelling software and BIM modelling processes are increasingly used in the Architecture, Engineering, and Construction (AEC) industry. In many countries, newly built buildings are modelled in BIM software, and delivered to the client as an as-built model. For existing buildings, techniques are available to semi-automatically create BIM models from 3D point cloud data of facilities, e.g.,~\cite{hichri2013point, bassier2020unsupervised}.
A dominant data standard in the AEC industry is the Industry Foundation Classes (IFC) \cite{IFC}. This standard focuses heavily on the interoperable exchange of 3D data across BIM authoring tools. An open and neutral IFC file can be exported from a BIM modelling tool, making semantic and 3D geometric data openly available (human- and machine-readable). Libraries are available to import and read data exported in IFC formats. 
In our approach, the building data exported in the IFC format is first parsed by the Python library IFCOpenShell which allows loading of the data, represented in the IFC file, in a Python environment maintaining its relations and attributes. The Algorithm to create the RDF graph is reported in Algorithm~\ref{alg:IFCConversion}. To construct the graph, we first create a list of all data entries of type \texttt{IfcSpace}, typically representing rooms, corridors, or (virtual) areas. The union of all spaces covers the entire area of the building. We then iterate over all elements of type \texttt{IfcSpace} to represent spatial relations at the semantic level. We rely on the IFC identifiers  \texttt{IfcBoundedby} and \texttt{IfcRelContainedInSpatialStructure} to create RDF triples to describe spatial relations, adjacency and containment respectively, between spaces and elements connected to them (see lines 1-10 in Algorithm~\ref{alg:IFCConversion}). We then iterate over all elements of type \texttt{IfcElement} to add their geometry in the form of 3D meshes, their material, and whether they are static or dynamic  (see lines 12-18 in Algorithm~\ref{alg:IFCConversion}). Meshes are encoded as a string, as a collection of faces, edges and vertices with coordinates relative to a local reference frame, and assigned as a property of their respective nodes. The transformation between the local reference frame and the global reference frame is also stored. The distinction between static and dynamic elements is done by comparing the IFC type of any element to a list describing the types to be considered static.

\begin{algorithm}
\SetAlgoNlRelativeSize{0}
\KwData{I (environmental model exported in IFC data format)}
\KwData{vSpace (vector of all elements of type IFCSpace)}
\KwData{vElement (vector of all elements of type IFCElement)}
\KwData{StaticTypes (list of element types to be considered static)}
\KwResult{K (graph world model)} 

\For{space\_i in vSpace}{
    add triple (space\_i, type, Space) to K \\
    \For{element in space.BoundedBy}
    {
        add triple (space\_i, adjacent, element) to K \\
    }
    \For{element in space.ContainsElements}
    {
        add triple (space\_i, contains, element) to K \\
    }%
}%
\For{For element in vElement}{%
    add triple (element, type, element.type) to K\\
    add triple (element, hasGeometry, element.3DMesh ) to K \\
    add triple (element, hasMaterial, element.Material) to K \\
    \If {element.type in StaticTypes}
    {
    add  triple  (element, isStatic, True ) to K \\
    }
    \Else {
    add  triple  (element, isStatic, False ) to K 
    }
    \For{For subelement in element.IsDecomposedBy}
    {
    add  triple  (element, hasSubElement, subelement) to K
    }
}
\caption{Algorithm to initialize the graph world model from an IFC description of the environment.}
\label{alg:IFCConversion}
\end{algorithm}
\section{Generation of skill-specific maps}\label{sec:map_generation}
\begin{figure}
    \centering
    \begin{subfigure}[b]{0.65\linewidth}
        \includegraphics[width=\columnwidth]{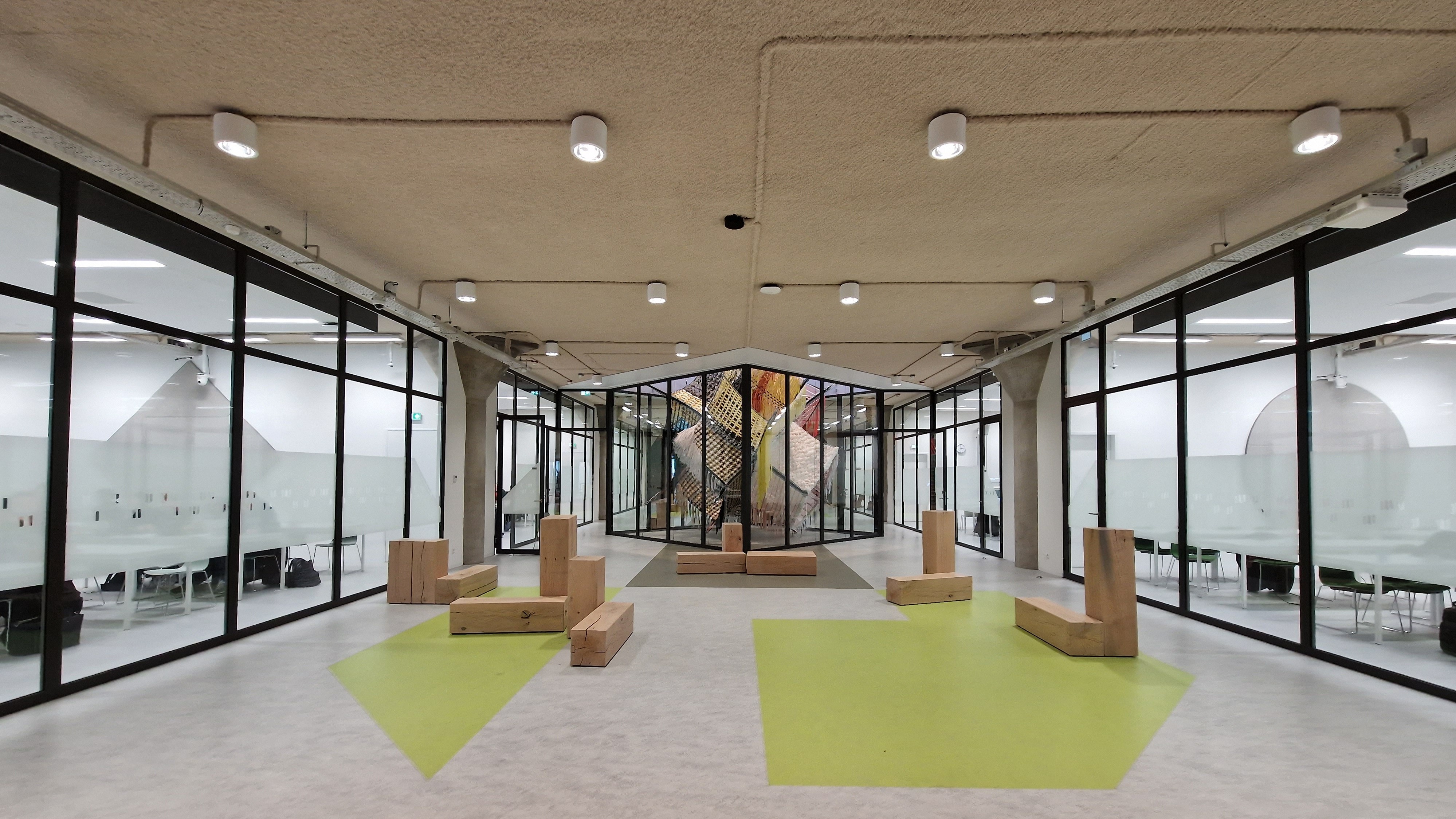}
    \end{subfigure}
    \caption{Photo of one of the spaces in the Neuron building at Eindhoven University of Technology with a large amount of glass elements.}
    \label{fig:neuron}
\end{figure}

In this section, we introduce methods to obtain skill-dependent maps from the graph world model derived by the procedure in Algorithm~\ref{alg:IFCConversion}. We specifically consider robots equipped with a 2D LIDAR sensor which are commonly used for both localization and navigation.
The algorithms proposed can interpret the data in the graph world model to generate 2D maps for navigation and localization. The elements to be represented in the maps can be selected based on their semantics such that the maps align with the specific localization or navigation skill. For the generation of a topological map from an RDF graph world model to facilitate high-level planning,  we refer to our previous work~\cite{de2021queries}. 
We validate the approach considering graph world models derived for two different environments.
The Neuron Building at the Eindhoven University of Technology in Figure~\ref{fig:neuron} and
secondly, a simplified lab environment consisting of two rooms with a connecting corridor in Figure~\ref{fig:simpleenviroment_physical}. Measures of the complexity and size of the models are reported in Table \ref{tab:modelcomplexity}. The simplified lab environment is used to validate the feasibility of integrating the maps with real-time control software of physical robots. To achieve this, we integrated the generated map with the MoveBase navigation framework in ROS~\cite{putz2018move}.

\begin{figure*}
\centering
  \begin{subfigure}{0.95\linewidth}
    \centering
    \includegraphics[width=\linewidth]{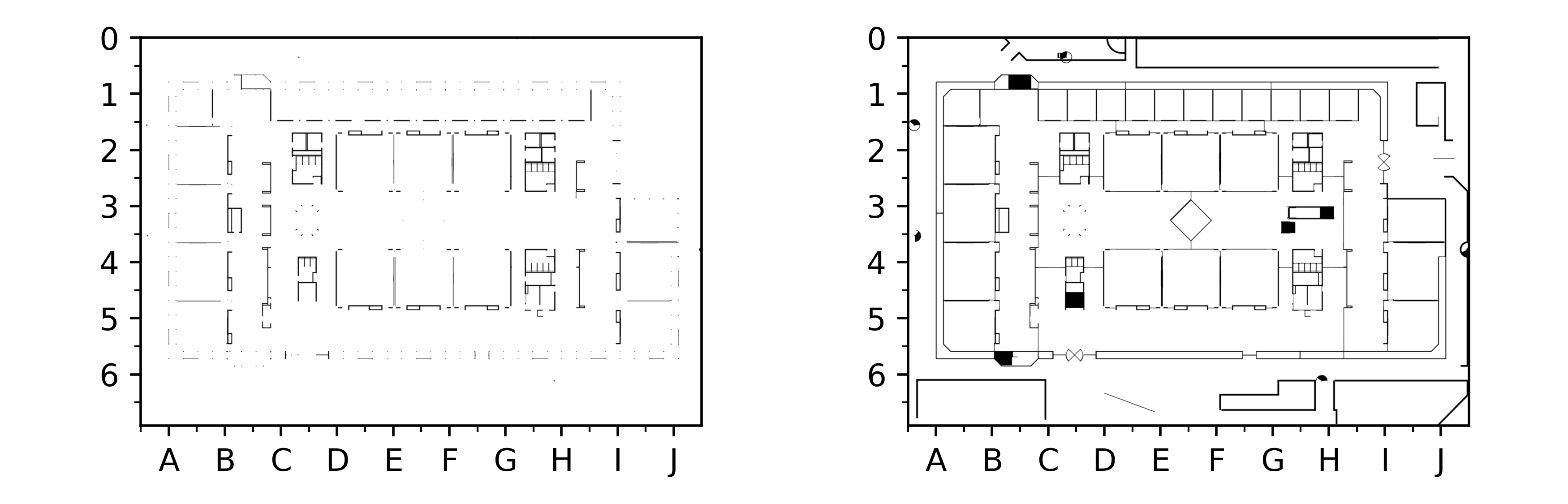}%
    \end{subfigure}
  ~
  \begin{subfigure}{0.95\linewidth}
    \centering
    {\includegraphics[width=\linewidth]{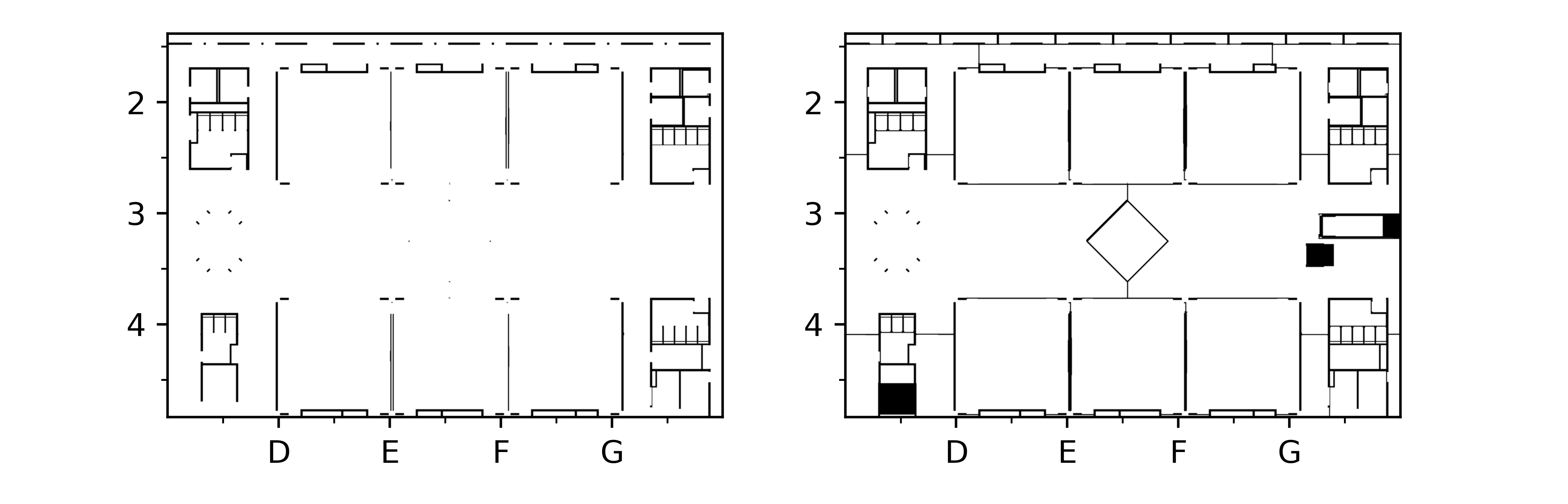}}
  \end{subfigure}%
  \caption{Generated occupancy grid maps of the Neuron building at Eindhoven University of Technology introduced in Figure~\ref{fig:neuron}. Note the absence of dynamic objects and furniture in the maps compared to the real world. Left column: Localization maps without glass elements. Right column: Navigation maps. Bottom row: Closeup of maps on the top row, which includes the space depicted in Figure \ref{fig:neuron}.}%
\label{fig:neuron0}%
\end{figure*}

\begin{figure}
    \centering
    \includegraphics[width=\linewidth]{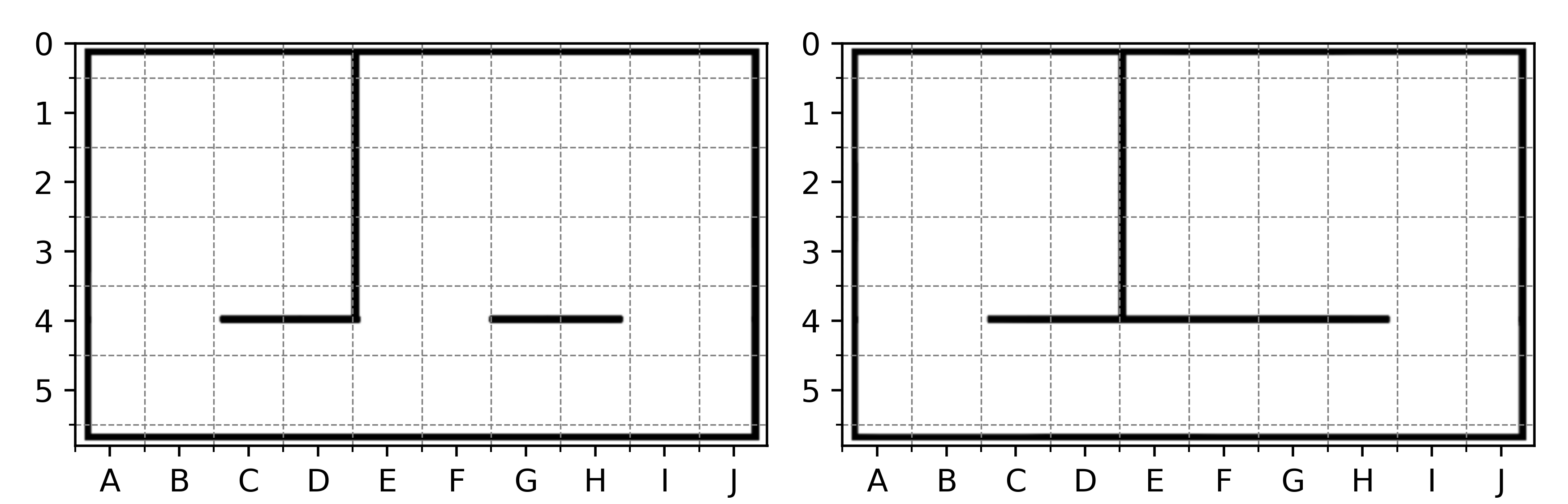}%
  \caption{Generated occupancy grid maps of the simplified lab setup introduced in Figure \ref{fig:simpleenviroment_physical} Left: Localization map. Right: Navigation map.}%
  \label{fig:SimpleEnv0}
\end{figure}

\subsection{2D maps for localization} 
The generated 2D maps for localization model the contour of the elements that would be seen by a planar LIDAR mounted at an arbitrary, but known, location on the robot. Since the graph describes the modeled environment in terms of individual components, it is possible to select a subset of the semantic elements for which we want to report the contour. This allows the generation of maps for semantic localization (e.g., Hendrikx et al.\cite{hendrikx2021connecting}) or state-of-practice localization based on particle filters (e.g., AMCL~\cite{fox2001kld}). The selection can be operated by SPARQL queries, providing a specification for whether an element should, or should not, be included, based on its type and (material) properties.
The procedure to generate the contour of the selected elements as seen by a planar LIDAR mounted at the given height, $h$, is provided in Algorithm~\ref{alg:localizationMap}.

\begin{figure}
  \centering
  \begin{subfigure}{0.475\columnwidth}
\includegraphics[width=\columnwidth]{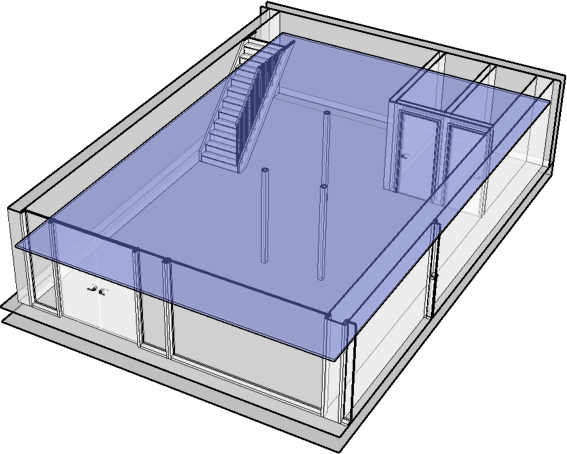}
  \end{subfigure}
  ~
  \begin{subfigure}{0.475\columnwidth}
  \includegraphics[width=\columnwidth]{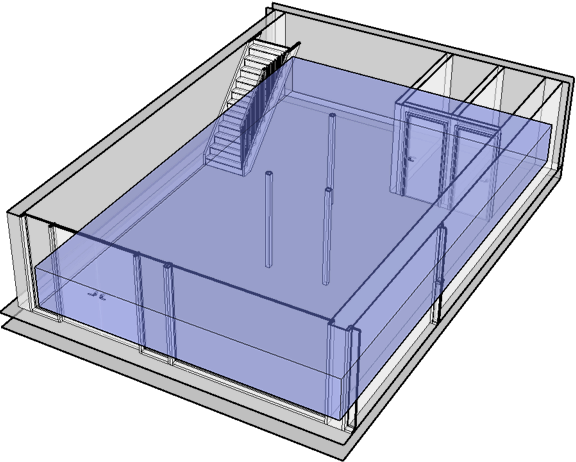}
  \end{subfigure}
  \caption{Representation of a plane (left) and volume (right), corresponding to the localization and navigation map generation respectively, intersecting the elements of a room in the House of Robotics model.}
  \label{fig:planeIntersection}
\end{figure}

The algorithm uses basic geometric computation techniques to compute the intersection between a solid rectangular cuboid $L$ representing the point of view of the LIDAR and the 3D mesh of each semantic element $M_i$ selected from the graph world model.
Given a mesh representation of $L$ and $M_i$, a point in Cartesian space $x$ and two membership functions $l(x)$ and $m(x)$ the Boolean intersection operation can be defined as
\begin{equation}
    L \cap M_i = \{ x \mid l(x) \text{ and } m_i(x) \}
\label{eq:intersection}
\end{equation}
The intersection points are then used to create a new mesh with related vertices, edges, and faces which correspond to the intersection area between $L$ and $M_i$. By representing the intersection area as occupied space in a binary grid map the intersection contour at the height of the LIDAR can be reported.
Efficient libraries are available to compute the intersection and the membership functions for polygonal meshes, such as the ones used in this paper. For this work, we rely on the geometry processing Python library Pymesh which is a wrapper around libGL for their computation.
A visualization of a cuboid intersecting the mesh of a wall and the resulting contour is visualized in Figure~\ref{fig:planeIntersection}. The full procedure is reported in Algorithm~\ref{alg:localizationMap}.

\begin{algorithm}
\KwData{K (graph world model)}
\KwData{$h_{lidar}$ (lidar height)}
\KwData{semanticElementTypes }
\KwResult{Localization map}
L = computePlane($h_{lidar}$)
\For{element in semanticElementTypes}{
    query K to retrieve  $M_i$\;
    compute $L \cap M_i$ (Eq.~\ref{eq:intersection})\;
    $I_i$ = mesh at intersection points \;
    \If{x in I }
    {
    draw black cell at x \;
    }
    
}%

\caption{Mesh Slicing and Polygon Construction Algorithm}
\label{alg:localizationMap}
\end{algorithm}

\subsubsection*{Validation}
We apply  Algorithm~\ref{alg:localizationMap} to the graph world model of the Neuron building (see Figure~\ref{fig:neuron0}) and to the model of the simplified lab setup (see Figure~\ref{fig:SimpleEnv0}).
For the Neuron building, we select a LIDAR height of $h=0.5$ [m] and for the lab enviroment a height of $h=0.1$ [m].
The list of \texttt{elementTypes} to be reported on the map for Neuron are all semantic element types except for \{\texttt{IFCOpeningElement}, \texttt{IfcFurnishingElement}, \texttt{IfcBuildingElementProxy}, \texttt{IfcFlowTerminal}\}. We furthermore filter out any element that is made of glass. For the simplified lab enviroment the list of elements considered is \texttt{semanticElementTypes = [IFCWall]}. The generated localization maps are displayed in the left column of Figures~\ref{fig:neuron0} and \ref{fig:SimpleEnv0}.

\subsection{2D navigation maps}
The process to generate a 2D navigation map is similar as the one to create a localization map. The generated maps represent as drivable the space which is not occupied by a mesh of any element at heights at or below that of the robot. The latter consideration accounts for changes in the floor-to-ceiling height due to, for example, items hanging from ceilings or structural constructions such as arches. The map is generated considering the intersection between a cuboid and the meshes of all elements known to the graph world model. The height of the cuboid is set at the height of the robot. The procedure is reported in Algorithm~\ref{alg:navigationMap}. 

\begin{algorithm}
\KwData{K (graph world model)}
\KwData{$h_{robot}$ (robot height)}
\KwData{semanticElementTypes }
\KwResult{Localization map}
L = computeCuboid($h_{robot}$)
    query K to retrieve  $M_i$\;
    compute $L \cap M_i$ (Eq.~\ref{eq:intersection})\;
    $I_i$ = mesh at intersection points \;
    \If{x in I }
    {
    draw black cell at x \;
    }
\caption{Algorithm to create the navigation map}
\label{alg:navigationMap}
\end{algorithm}
\subsubsection*{Results}
We apply  Algorithm~\ref{alg:navigationMap} to the graph world model of the Neuron building (see Figure~\ref{fig:neuron0}) and to the model of the simplified lab setup (see Figure~\ref{fig:SimpleEnv0}).
For the Neuron building, we select a robot height of $h=1.5$ [m] while for the simplified lab setup a height of $h=0.5$ [m]. The list of \texttt{elementTypes} to be reported on this map is the full set of elements. For the lab environment, the list of elements considered is \texttt{semanticElementTypes = [IFCWall]}. The generated navigation maps are displayed in the right column of Figures~\ref{fig:neuron0} and \ref{fig:SimpleEnv0}. Compared to the earlier generated localization maps, clear differences can be observed. For example, in the maps generated for the Neuron building in Figure \ref{fig:neuron0} it is clear that a large amount of the structural elements are made of glass. Hence the localization map contains a smaller amount of occupied cells, as the maps were generated under the assumption that the 2D lidar sensor would be unable to detect these elements. In Figure \ref{fig:SimpleEnv0} no glass elements are present, however, the arch in the model is occupied in the navigation map, but not in the localization map. This is due to the mismatch between the robot height and the height of the lidar sensor. 

\subsection{Integration}
\begin{figure}
  \centering
  \includegraphics[width=\columnwidth]{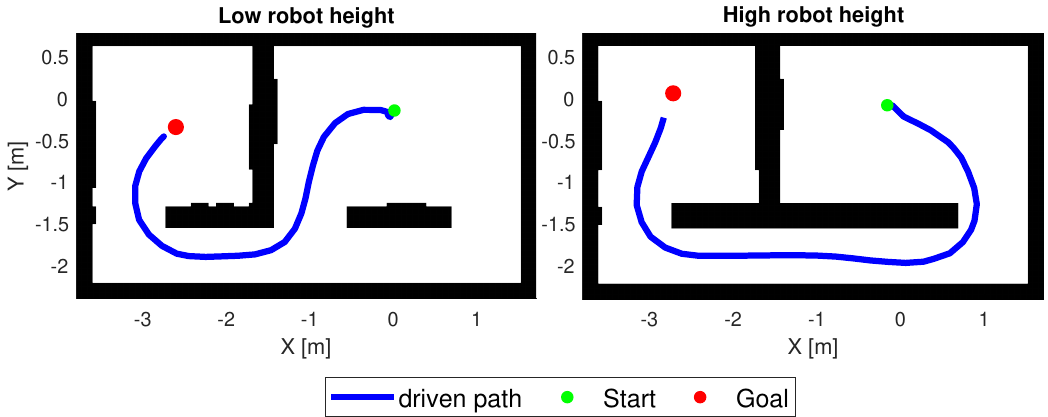}
  \caption{Traversed path during the physical validation experiments. Left: The path of a robot with a height of 0.25 [m]. Right: The path of a robot with a height of 0.5 [m]}
  \label{fig:TraversedPath}
\end{figure}

To further validate the approach and its suitability for integration with state-of-practice navigation frameworks, we first generate localization and navigation maps for the simple environment described in Figure~\ref{fig:SimpleEnv0}. The generated maps are then imported into the ROS parameter server and connected to the navigation framework MoveBase~\cite{putz2018move}.
MoveBase uses A* for global path planning based on the generated navigation map, a Dynamic Window Approach to sample local paths for obstacle avoidance~\cite{putz2018move} and a particle filter based localization performed on the generated navigation map using the AMCL framework \cite{fox2001kld}.
To show how localization and navigation maps can potentially differ, we consider a robot height of $h=0.5$ [m], and a sensor height of $h=0.1$ [m]. Considering this height, the robot will not be able to pass under the arch which should be reflected in the generated navigation map.
Integration tests were carried out with a ROSbot platform. The physical environment and the ROSbot are visible in Figure~\ref{fig:simpleenviroment_physical}.
The generated localization and navigation maps, obtained by applying Algorithm~\ref{alg:localizationMap} and Algorithm~\ref{alg:navigationMap} to the graph world model of the environment are visualized in Figure~\ref{fig:SimpleEnv0}.
The robot is tasked to move from point A to point B. The traversed trajectory is plotted in Figure~\ref{fig:TraversedPath}, for both the robot height cases. Please note that the tall robot's path stays away from the arch. A video of the physical robot navigating from point A to point B is available.\footnote{\url{https://www.youtube.com/watch?v=R2rFwQd0PEY}}

\begin{table}[]
\caption{Summary on the complexity and element numbers and file size of the considered BIM models and their respective graph world models.}
\centering
\begin{tabular}{l|ll}
\hline
                   & Model           &  \\ \hline
Lines in IFC file & Neuron Building & 3.8e6  \\
                   & Lab environment  & 651    \\ \hline
Number of Triples  & Neuron Building & 2.4e5  \\
                   & Lab environment  & 221   \\ \hline
.ttl file size [MB]& Neuron Building & 130\\
                   & Lab environment  & 4e-2 \\\hline
Number of Elements & Neuron Building & 11500 \\
                   & Lab environment  & 11   \\ \hline
Number of Meshes   & Neuron Building & 10822\\
                   & Lab environment  & 11  \\ \hline
\end{tabular}
\label{tab:modelcomplexity}
\end{table}

\section{Discussion and conclusion}\label{sec:DiscAndConcl}
This paper presented methods to create graph world models of environments which are initialized from BIM models and from which robot-specific maps can be extracted.
The paper validated the presented methods on a graph world model representing a large and complex university building as well as a lab setup. The robot-specific maps focused on robots equipped with 2D LIDAR sensors used for localization and navigation. Results show the potential of the approach in terms of the automatic process of graph initialization and map generation.\\
Enabling the generation of maps for other types of sensors such as 3D LIDARs or RGB-D cameras represents an interesting development of the work presented in this paper. 
Additionally, the maps generated from the graph world models might be incomplete, for example, furniture might not be represented at initialization but might need to be represented in the maps. Such considerations call for further research into the update of the graph world model based on live data gathered by (heterogenous) robots. We anticipate that merging of information from different robots will be facilitated by the fact that they can ground their sensory observation on maps generated from the same source which, by design, are aligned in terms of definition of reference frames. Yet, determining when and how to update the shared graph world model remains an interesting next step to further develop the work presented in the paper.

\bibliographystyle{IEEEtran}
\bibliography{bibliography}

\end{document}